\documentclass[final]{cvpr}

\usepackage{times}
\usepackage{epsfig}
\usepackage{graphicx}
\usepackage{amsmath}
\usepackage{amssymb}
\usepackage{subfigure}

\usepackage[pagebackref=true,breaklinks=true,colorlinks,bookmarks=false]{hyperref}

\def\cvprPaperID{5189} 
\def\confYear{CVPR 2021}
\begin{document}

\title{Sketch2Model: View-Aware 3D Modeling from Single Free-Hand Sketches}

\author{Song-Hai Zhang\thanks{corresponding author.} \quad Yuan-Chen Guo \quad Qing-Wen Gu\\
BNRist, Department of Computer Science and Technology, Tsinghua University, Beijing\\
{\tt\small shz@tsinghua.edu.cn,\;guoyc19@mails.tsinghua.edu.cn,\;gqw17@mails.tsinghua.edu.cn}
}

\maketitle

\begin{abstract}
  We investigate the problem of generating 3D meshes from single free-hand sketches, aiming at fast 3D modeling for novice users. It can be regarded as a single-view reconstruction problem, but with unique challenges, brought by the variation and conciseness of sketches. Ambiguities in poorly-drawn sketches could make it hard to determine how the sketched object is posed. In this paper, we address the importance of viewpoint specification for overcoming such ambiguities, and propose a novel view-aware generation approach. By explicitly conditioning the generation process on a given viewpoint, our method can generate plausible shapes automatically with predicted viewpoints, or with specified viewpoints to help users better express their intentions. Extensive evaluations on various datasets demonstrate the effectiveness of our view-aware design in solving sketch ambiguities and improving reconstruction quality. 
\end{abstract}

\section{Introduction}
\begin{figure}
\begin{center}
\includegraphics[width=0.99\linewidth]{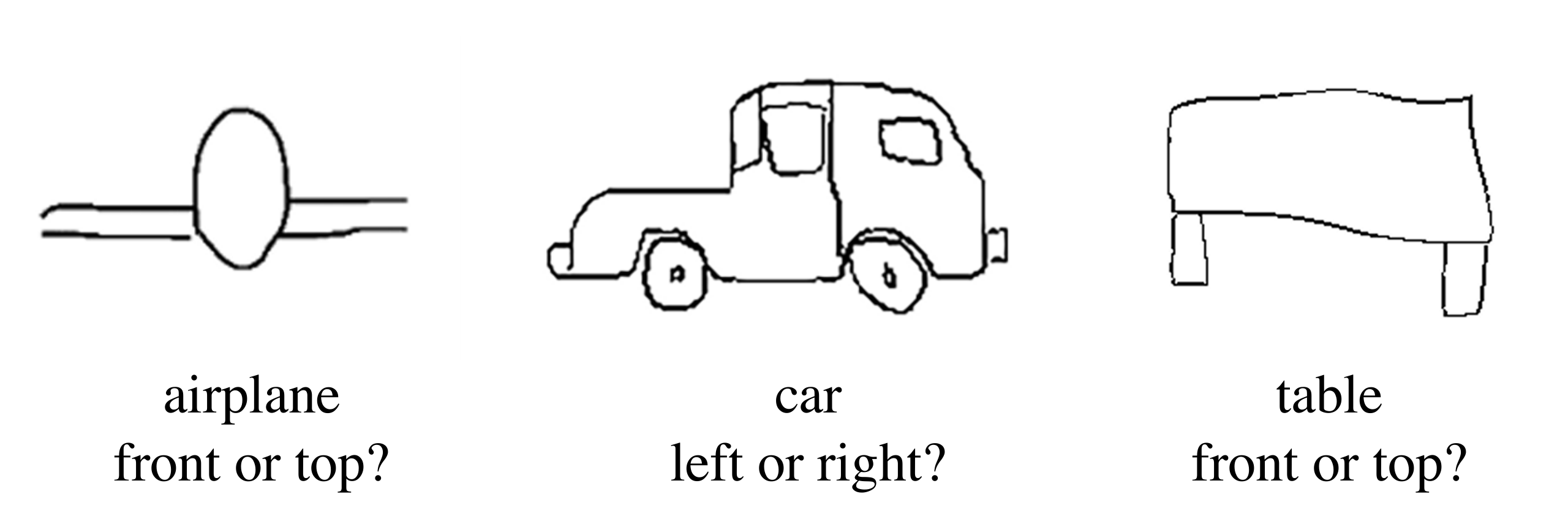}
\end{center}
   \caption{Ambiguities in free-hand sketches. It could be hard to determine how the sketched object is posed.}
\label{fig:sketch_ambiguity}
\end{figure}

Sketch-based 3D modeling has been studied for decades, intended to free people from the tedious and time-consuming modeling process. With the widespread usage of portable touch screens and emergence of VR/AR technologies, the need of 3D content creation for novice users is increasing~\cite{wang2020vr}. However, existing sketch-based 3D modeling techniques either require precise line-drawings from multiple views or simply retrieve from existing models, both failing to provide easy-to-use interface and customizability at the same time. We aim at fast and intuitive 3D modeling for people without professional drawing skills, and investigate the problem of mesh generation from a single free-hand sketch. It is in general a single-view reconstruction problem, but has its unique challenges due to the characteristics of free-hand sketches. Unlike real images, free-hand sketches can be poorly-drawn with drastic simplifications and geometric distortions, and lack important visual cues like textures or shading. This makes it difficult, sometimes even impossible, for algorithms to understand sketches correctly. One tricky scenario results from the ``camera-shape ambiguity''~\cite{li2020self}, as shown in Fig.~\ref{fig:sketch_ambiguity}. Subjectivity of sketch creation could lead to different explanation of the same sketch, like which side being the front of the car (left). Furthermore, a recent study shows that existing single-view reconstruction approaches mainly rely on the recognition of input images, rather than actually performing geometry reconstruction. Therefore, they can be hard to generalize to unseen data and perform worse on hand-drawn sketches, which are created with barely any constraint, showing more diverse in appearance.

Recall how human cognitive system works for the task of 3D reconstruction. Human relies on visual memory and visual rules to recover 3D geometry from imagery~\cite{olsen2009sketch}. Visual memory contains priors gained from daily life, like what an object of a certain class might look like, while visual rules act like a regularizer, requiring that an object should match what it looks like from a certain viewpoint. When seeing a sketch, we search our visual memory and try to find a solution that also obeys visual rules. However, this search process may not go well when (1) we could not figure out how the sketched object is posed, thus finding difficulty applying visual rules; (2) multiple solutions exist due to the ambiguity of the sketch. Both problems result from the lack of details in sketches, and we address that viewpoint specification can be essential to the above problems. If we are told which viewpoint the object is sketched from, search space can be easily reduced, which lowers search difficulty and may eliminate ambiguities.

Based on the above observations, we think that explicitly specified viewpoints can directly guide the reconstruction algorithm to find the correct shape described by the user. Moreover, viewpoints offer extra information for the understanding of hand-drawn sketches, which can improve generalization ability for out-of-distribution data. To utilize view information, we propose a view-aware sketch-based 3D reconstruction method, which explicitly conditions the generation process on a given viewpoint. We extend the traditional encoder-decoder architecture used in some recent works~\cite{kato2018neural,liu2019soft}. In the encoder, image features are decomposed to a latent view code and a latent shape code. The latent view code is used to predict viewpoint for the input sketch, and the latent shape code is combined with a latent view code to generate the output shape. Just like other disentangling problems, the network would take a short path and fail to decompose certain properties if no other constraints are made. To ensure that only view information is contained in the latent view space, a view auto-encoder is learned to transform viewpoints to latent view space and vice versa. To force the decoder network to depend on the given view code, we propose a simple but effective random view reinforced training strategy, where random views along with ground truth views are fed to the decoder, forcing the network to learn how to match the input sketch from different viewpoints. We use synthetic sketch data for training, due to the lack of paired sketch-3D dataset, and apply domain adaptation technique to bridge the gap between synthetic sketches and real sketches. In the inference stage, our method can generate 3D mesh based on a single free-hand sketch automatically using the predicted viewpoint, or semi-automatically using a user-specified viewpoint. 

Extensive experiments are conducted to demonstrate the effectiveness of our design. We first perform case studies to show characteristics of our view-aware architecture, and how specified views can be used to eliminate ambiguities. To evaluate how our method performs on free-hand sketches and inspire further research, we collect a ShapeNet-Sketch dataset based on the ShapeNet dataset, which contains 1,300 sketch-shape pairs. Results on various datasets show that our method can generate higher-quality shapes and better convey user's intentions comparing to alternative baselines, which demonstrates the advantages of introducing viewpoint specification. Ablation studies are performed to show the necessity of serveral parts of our method.

To summarize, our contributions are as follows:
\begin{itemize}
    \item We are the first to investigate the problem of generating 3D mesh from a single free-hand sketch, which provides a fast and easy-to-use 3D content creation solution for novice users.
    \item We address the importance of viewpoint specification in the sketch-based reconstruction task, and design a view-aware generation architecture to condition the generation process explicitly on viewpoints. Quantitative and qualitative evaluations on various datasets demonstrate our method can generate promising shapes that well convey user's intentions and generalize better on free-hand sketches.
\end{itemize}

\begin{figure*}[t]
\begin{center}
\includegraphics[width=0.99\linewidth]{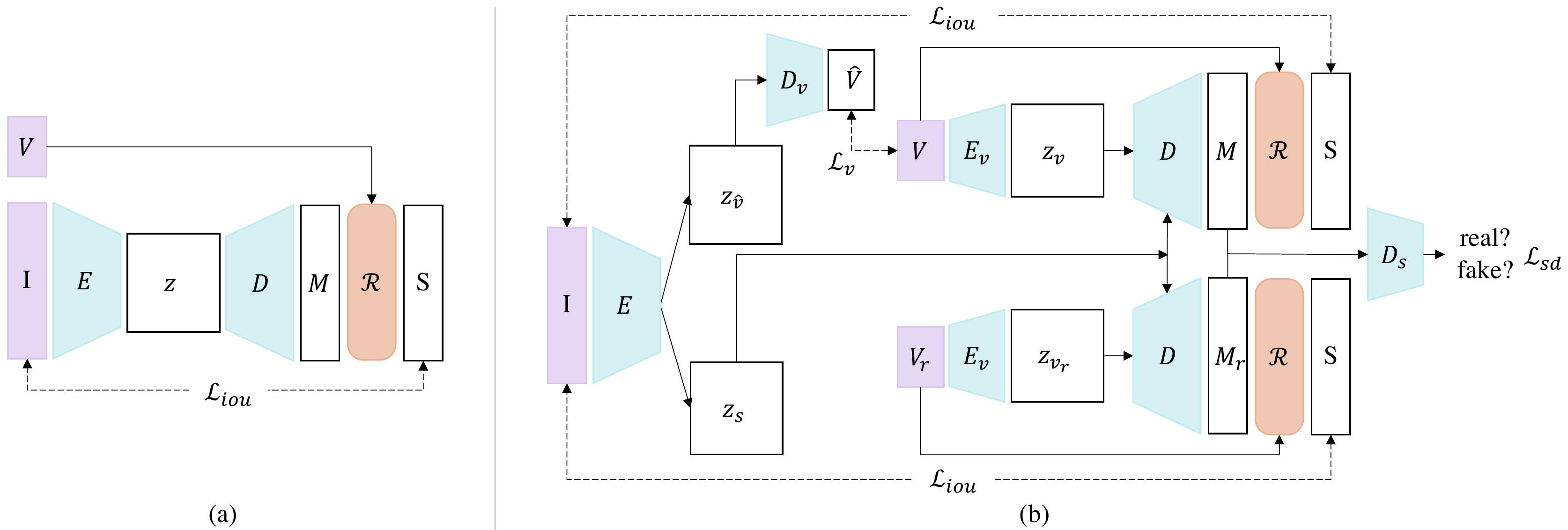}
\end{center}
   \caption{Network architecture of (a) baseline methods~\cite{kato2018neural,liu2019soft} and (b) our method. Purple and blue blocks denote inputs and trainable parameters respectively. $\mathcal{R}$ is the differentiable rendering module. Our method extends the encoder-decoder model by decomposing image features into shape space $\mathcal{S}$ and view space $\mathcal{V}$. We train a view encoder $E_v$ and a view decoder $D_v$ to ensure only view-related information is contained in view space, and force the decoder $D$ to use view information by training with an additional random view $V_r$. Shape quality is preserved by a shape discriminator $D_s$.}
\label{fig:network}
\end{figure*}

\section{Related Work}
\subsection{Single-View 3D Reconstruction}
Recovering 3D geometry from a single image is an ill-posed problem. Early approaches utilize perceptual cues, such as shading~\cite{horn1970shape} and texture~\cite{witkin1981recovering}, to get a clue about surface orientations. With the emergence of online 3D model collections and large-scale 3D model datasets~\cite{chang2015shapenet}, data-driven approaches are developed to infer category-specific shapes directly from image features~\cite{huang2015single,choy20163d,girdhar2016learning,fan2017point,wang2018pixel2mesh,mescheder2019occupancy,chen2019learning,pontes2018image2mesh}, in the form of voxel~\cite{choy20163d,girdhar2016learning}, point cloud~\cite{fan2017point}, triangle mesh~\cite{huang2015single,wang2018pixel2mesh,pontes2018image2mesh}, and implicit function~\cite{mescheder2019occupancy,chen2019learning}. Recently, differentiable rendering techniques~\cite{yan2016perspective,tulsiani2017multi,kato2018neural,liu2019soft,niemeyer2020differentiable} are proposed to relax the need for ground truth 3D models in training, which enables supervision with solely 2D images. Our method is based on the single-view reconstruction process of~\cite{liu2019soft}. We extend the original approach by disentangling image features into a latent shape space and a latent view space, and achieve view-aware generation by applying special designed training strategy. Note that no 3D information is needed in our training process.
\subsection{Sketch-Based 3D Modeling}
3D modeling based on sketches, has been studied for a long time. Olsen \etal~\cite{olsen2009sketch} categorized traditional sketch-based modeling techniques into evocative~\cite{funkhouser2003search,chen2003visual,wang2015sketch} and constructive~\cite{igarashi2006teddy,nealen2007fibermesh,karpenko2006smoothsketch,miao2015symmsketch} ones. Evocative approaches create models from template primitives or retrieve objects from model collections, while constructive ones directly map the input strokes to a 3D model, providing free-form modeling abilities. Either of the two paradigms has its shortcomings: evocative methods are confined to generate certain shapes or primitives, and the lack of prior knowledge in constructive methods often requires large efforts to get the desired shape. Deng \etal~\cite{deng2019interactive} proposed to extract free-form shapes from single images in an interactive way, inspired by the lofting technique in CAD systems. By utilizing deep learning techniques, several works~\cite{lun20173d,huang2016shape,han2017deepsketch2face,su2018interactive,delanoy20183d,wang2018unsupervised,shen2019deepsketchhair,wang20203d,han2020reconstructing} have achieved to gain priors from existing 3D models, while being able to leverage user inputs. However, very few of them focuses on reconstruction from free-hand sketches~\cite{wang2018unsupervised,wang20203d}, instead of precise line drawings. Wang \etal~\cite{wang2018unsupervised} proposed a retrieval-based apporach to generate 3D voxels from free-hand sketches. The work most similar to ours is that of Wang \etal~\cite{wang20203d}, where point cloud is generated according to a single free-hand sketch, and rotated to a canonical view by predicted viewpoint to match the ground truth. Different from their work, we regard viewpoint as an additional input to the generation process, and rely on them to interpret sketches. The authors proposed sketch standardization steps to synthesize sketches from edge maps, while we try to bridge the domain gap in feature space.

\section{Method}
To generate 3D mesh from a single free-hand sketch, we first propose a baseline approach, that works the same way as single-view reconstruction for real images. Then we extend the baseline architecture by decomposing image features into a latent view space and a latent shape space, and condition the generation process on the choice of viewpoint.

\subsection{View-Aware Generation}
\label{sec:view-aware}
Given an input sketch $I$, we wish to generate a 3D mesh $M$, under the constraint that the sketch represents the object seen from a specific viewpoint $V$.

As a basis of our method, the single-view mesh reconstruction framework proposed in \cite{kato2018neural,liu2019soft} is built upon a simple encoder-decoder architecture, shown in Fig.\ref{fig:network}(a). The encoder $E$ is a convolutional neural network, which extracts image features from input image $I$ and outputs a compact feature vector $z=E(I)$. The decoder $D$ calculates from $z$ the vertex offsets of a template mesh, and deforms it to get the output mesh $M=D(z)$. By approximating gradients of the rasterization process in the differentiable rendering module $\mathcal{R}$, the network can be end-to-end trained by comparing rendered silhouettes and ground truth ones, without the need for ground truth 3D shapes.

To explicitly condition the generation process on viewpoints, we design a novel view-aware architecture, shown in Fig.\ref{fig:network}(b). In our encoder, the image feature vector $z$ is further mapped to a latent view space $\mathcal{V}$ and a latent shape space $\mathcal{S}$. The decoder takes a latent view code $z_v$ and a shape code $z_{s}$, and get a deformed mesh $M$. Denote the image space and mesh space to be $\mathcal{I}$ and $\mathcal{M}$, our encoder and decoder can be expressed as:
\begin{align}
    E&:\mathcal{I}\rightarrow\mathcal{V}\times\mathcal{S}\\
    D&:\mathcal{V}\times\mathcal{S}\rightarrow\mathcal{M}
\end{align}

We also introduce a view auto-encoder, denoted as $E_v$ and $D_v$, to encode an arbitrary viewpoint to the latent view space and vice versa:
\begin{align}
    D_v(E_v(V))&=V\\
    E_v(V)&\in\mathcal{V}
\end{align}
such that only view-related information are held in the latent view space. We predict the viewpoint for the input sketch from the latent view code $z_{v}$ given by the encoder:
\begin{equation}
    \hat{V}=D_v(z_v)
\end{equation}

Intuitive as it seems, it is not straightforward to ensure that shapes are generated by taking the latent view code $z_v$ into consideration. A common degradation could happen, that $M$ is directly generated from $z_{s}$, totally ignoring $z_v$, if trained without any other constraint. We propose a random view reinforced training strategy to solve this problem. In addition to the ground truth viewpoint $V$, we feed a random viewpoint $V_r$ to the decoder network and generate a mesh $M_r$. By forcing consistency between the projected silhouette of $M_r$ under $V_r$ and the input silhouette, the network will learn to match the input sketch under any given viewpoint. However, interpreting input sketches with an arbitrary viewpoint is not feasible. This will cause the network to generate distorted shapes, which only satisfy the silhouette constraint, but are not meaningful objects. To avoid distortions as much as possible, we introduce a shape discriminator $D_s$, which is trained jointly with the encoder and decoder in an adversarial manner. The random view reinforced training and the shape discriminator work as a balance between view-awareness and shape quality. The network may not produce shapes strictly matching the input silhouette under $V_r$, but still be aware of the impacts of viewpoints to the generation results. 
\subsection{Training and Inference}
\paragraph{Domain adaptation.} Due to the lack of paired sketch-shape data, we use synthetic sketches of ShapeNet objects as training data, which are precise line-drawings of objects, and have different appearances from free-hand sketches. Performance drop occurs when the model trained on synthetic sketches is tested on hand-drawn ones. To bridge the gap between them, previous works mostly try to mimic hand-drawn sketches when generating synthetic data by hand-crafted rules or image translation techniques. We first propose to bridge this gap in feature space instead of image space, inspired by existing domain adaptation approaches. Our intention is to make image features of synthetic and hand-drawn sketches indistinguishable. To do this, we learn a domain discriminator to separate features coming from the two domains, and train it with the encoder adversarially.

\paragraph{Optimization objectives.} Our loss functions consist of five parts: silhouette loss $\mathcal{L}_s$ between projected and ground truth silhouettes, geometry regularizations $\mathcal{L}_r$ for generated shapes, view prediction loss $\mathcal{L}_v$ between predicted and ground truth viewpoints, classification loss $\mathcal{L}_{sd}$ and $\mathcal{L}_{dd}$ for shape discriminator and domain discriminator. 

For input and rasterized silhouettes, we adopt IoU loss as in \cite{kato2018neural,liu2019soft}. Let $S_1$ and $S_2$ be two binary silhouettes, IoU loss $\mathcal{L}_{iou}$ is defined as:
\begin{equation}
    \mathcal{L}_{iou}(S_1,S_2)=1-\frac{||S_1\otimes S_2||_1}{||S_1\oplus S_2-S_1\otimes S_2||_1} \end{equation}
Denote $P(\cdot,\cdot)$ as the differentiable rasterization function, which takes a mesh $M$ and a viewpoint $V$, and output the rasterized silhouette of $M$ under $V$. Our silhouette loss is expressed as:
\begin{equation}
\begin{split}
    \mathcal{L}_s &= \mathcal{L}_{iou}(S,P(D(z_{s},E_v(V)),V))\\
    &+ \lambda_{r}\mathcal{L}_{iou}(S,P(D(z_{s},E_v(V_r)),V_r))
\end{split}
\end{equation}
where $V$ and $S$ is the ground truth viewpoint and silhouette respectively. Note that we use the latent code for ground truth viewpoint $E_v(V)$ in training, instead of the predicted $z_v$, because we observe negative impacts for training by utilizing inaccurate predictions. Most existing works compute IoU loss on low resolution silhouettes like $64\times64$, which ignores fine structures. Considering computational efficiency, we adopt a multi-scale progressive training strategy, leveraging silhouettes of different resolutions. Let $L_{s}^{i}$ be the silhouette loss on the i-th level of the silhouette pyramid, $N$ be the number of levels, we adopt a progressive silhouette loss $\mathcal{L}_{sp}$:
\begin{equation}
    \mathcal{L}_{sp}=\sum_{i=1}^{N}\lambda_{s_{i}}\mathcal{L}_{s}^{i}
\end{equation}
In early stage of training, only low resolution silhouettes are used for loss calculation. Higher resolution ones are used for shape refinement as the training progresses. 

To improve visual quality of generated meshes, we apply flatten loss and and laplacian loss as regularizations for mesh structures, as in~\cite{kato2018neural,liu2019soft,wang2018pixel2mesh}. We denote the shape regularization loss as $\mathcal{L}_{r}$.

For the representation of viewpoint, we assume fixed distance to camera and represent viewpoints as Euler angles. We adopt L2 loss for predicted and ground truth views:
\begin{equation}
    \mathcal{L}_{v}=||V-\hat{V}||_2=||V-D_v(z_v)||_2
\end{equation}
To train the view encoder, we adopt a view reconstruction loss $\mathcal{L}_{vr}$ between original and reconstructed views:
\begin{align}
    \mathcal{L}_{vr}=||V-D_v(E_v(V))||_2
\end{align}

\begin{figure*}
\begin{center}
\includegraphics[width=0.99\linewidth]{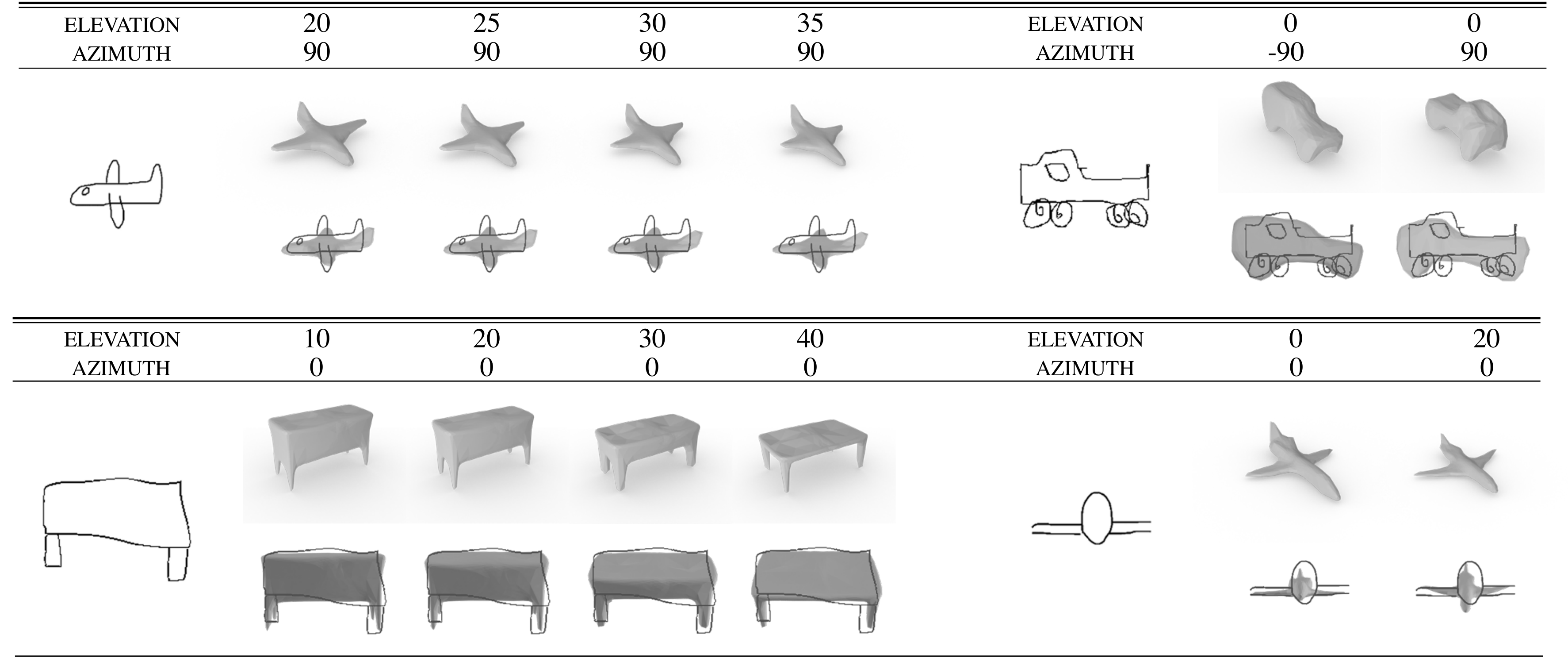}
\end{center}
   \caption{Illustrations for effectiveness of our view-aware design. When giving the same sketch as input but specifying different viewpoints, our model can synthesize different results trying to match the input shape from the corresponding point of view. For each example, the first row shows synthesized results from a fixed viewpoint, and the second row shows projections onto the given viewpoint overlapped with the input sketch. The fine property of our view-aware design allows solving ambiguity problems with minimal user input.}
\label{fig:view_aware}
\end{figure*}

To ease the training process, we apply a gradient reversal layer before shape discriminator and domain discriminator. The gradient reversal layer reverses the sign of the gradient that passes it, which achieves adversarial training in a single pass. For the shape discriminator, we expect it to output 1 for $M$ and 0 for $M_r$, and for the domain discriminator, 1 for synthetic sketches and 0 for hand-drawn sketches, which lead to the following losses:
\begin{equation}
    \begin{split}
    \mathcal{L}_{sd}=&-\mathbb{E}_{M}[log(D_s(M))]\\
    &-\mathbb{E}_{M_r}[1-log(D_s(M_r))]
    \end{split}
\end{equation}
\begin{equation}
    \begin{split}
    \mathcal{L}_{dd}=&-\mathbb{E}_{S_s}[log(D_d(S_s))]\\
    &-\mathbb{E}_{S_h}[1-log(D_d(S_h))]
    \end{split}
\end{equation}
where $S_s$ and $S_h$ denote synthetic sketches and hand-drawn sketches respectively. 

Our overall loss function is the weighted sum of the above objectives:
\begin{equation}
\begin{split}
    \mathcal{L}&=\mathcal{L}_{sp}+\mathcal{L}_{r}+\lambda_{v}\mathcal{L}_{v}+\lambda_{vr}\mathcal{L}_{vr}\\
    &+\lambda_{sd}\mathcal{L}_{sd}+\lambda_{dd}\mathcal{L}_{dd}
\end{split}
\end{equation}

During inference, shapes can be generated from an input free-hand sketch automatically or semi-automatically, depending on the choice of latent view code $z_v$. Users can use the predicted viewpoint $\hat{V}$ to get the output mesh $\hat{M}=D(z_s, E_v(\hat{V}))$,
or provide a specified viewpoint $\tilde{V}$ to get $\tilde{M}=D(z_s, E_v(\tilde{V}))$.

\begin{table*}[h]
\footnotesize
\begin{center}
\renewcommand\tabcolsep{4.3pt}
\begin{tabular}{c||ccccccccccccc|c}
\hline
 & \multicolumn{14}{c}{Voxel IoU ($\uparrow$)} \\
Category & airplane & bench & cabinet & car & chair & display & lamp & loudspeaker & rifle & sofa & table & telephone & watercraft & mean \\\hline\hline
Retrieval & 0.513 & 0.380 & 0.518 & 0.667 & 0.346 & 0.385 & 0.325 & 0.468 & 0.475 & 0.483 & 0.311 & 0.622 & 0.422 & 0.455\\
SoftRas & 0.576 & 0.467 & 0.663 & \textbf{0.769} & 0.496 & 0.541 & 0.431 & 0.629 & 0.605 & 0.613 & \textbf{0.512} & 0.706 & 0.556 & 0.582\\\hline
Ours (pred) & 0.618 & 0.477 & 0.667 & 0.746 & 0.515 & 0.550 & 0.463 & 0.624 & 0.606 & 0.620 & 0.470 & 0.673 & 0.569 & 0.584\\
Ours (gt) & \textbf{0.624} & \textbf{0.481} & \textbf{0.701} & 0.751 & \textbf{0.522} & \textbf{0.604} & \textbf{0.472} & \textbf{0.641} & \textbf{0.612} & \textbf{0.622} & 0.478 & \textbf{0.719} & \textbf{0.586} & \textbf{0.601}\\
\hline
 & \multicolumn{14}{c}{Chamfer Distance ($\downarrow$)} \\
Category & airplane & bench & cabinet & car & chair & display & lamp & loudspeaker & rifle & sofa & table & telephone & watercraft & mean \\\hline\hline
Retrieval & 0.856 & 1.384 & 1.941 & \textbf{0.767} & 1.878 & 1.967 & \textbf{3.017} & 2.468 & 0.731 & 2.056 & 2.151 & 1.042 & 1.130 & 1.645\\
SoftRas & 0.531 & \textbf{0.944} & 2.271 & 0.947 & 1.703 & 1.762 & 3.272 & 2.234 & \textbf{0.495} & 2.161 & \textbf{1.662} & 1.307 & \textbf{0.872} & 1.551\\\hline
Ours (pred) & 0.493 & 1.003 & 1.867 & 0.812 & 1.488 & 1.637 & 3.300 & 2.021 & 0.604 & \textbf{1.944} & 1.790 & 1.120 & 1.018 & 1.469\\
Ours (gt) & \textbf{0.470} & 0.974 & \textbf{1.561} & 0.790 & \textbf{1.457} & \textbf{1.228} & 3.263 & \textbf{1.884} & 0.574 & 2.011 & 1.744 & \textbf{0.928} & 0.897 & \textbf{1.368}\\
\hline
\end{tabular}
\end{center}
\caption{Comparisons of mean voxel IoU and Chamfer Distance on ShapeNet-Synthetic test set. Our method with either ground truth viewpoins (gt) or predicted viewpoints (pred) surpasses baseline methods on most of the classes on voxel IoU score, and achieves the best overall performance regarding to Chamfer Distance. }
\label{tab:compare_shapenet}
\end{table*}

\begin{figure}
\begin{center}
\includegraphics[width=0.99\linewidth]{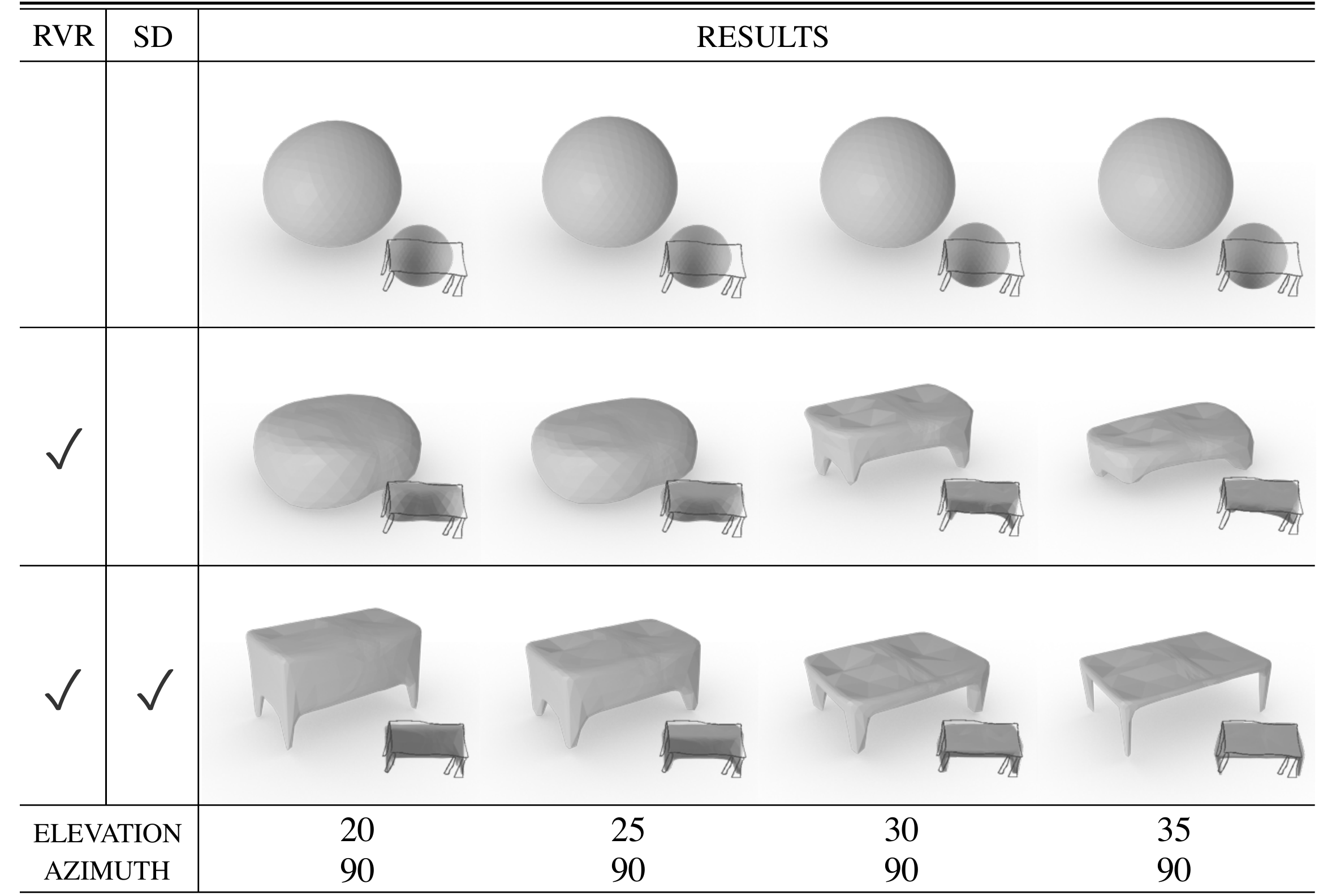}
\end{center}
   \caption{Ablation studies for random view reinforced training strategy (RVR) and shape discriminator (SD). Without applying random view reinforced training, the network would fail to generate meaningful objects under given viewpoints (first row). Without shape discriminator, the output shape may not resemble real objects (second row). We successfully achieve view-aware generation by adopting both techniques (third row).}
\label{fig:view_aware_ablation}
\end{figure}
\begin{table*}[h]
\footnotesize
\begin{center}
\begin{tabular}{ccc||ccccccc}
\hline
 RVR & SD & MS & airplane & bench & cabinet & car & chair & display & lamp \\\hline\hline
& & & 0.557 (0.565) & 0.345 (0.460) & 0.597 (0.579) & 0.747 (\textbf{0.753}) & 0.499 (0.508) & 0.457 (\textbf{0.577}) & 0.290 (0.421)\\
\checkmark & & & 0.587 (0.469) & 0.407 (0.435) & 0.674 (0.641) & 0.753 (0.736) & 0.514 (0.505) & 0.525 (0.504) & 0.454 (0.451)\\
\checkmark & \checkmark & & 0.588 (0.582) & 0.459 (0.455) & 0.690 (0.658) & \textbf{0.756} (0.749) & 0.510 (0.504) & 0.591 (0.540) & 0.461 (0.453)\\
\checkmark & \checkmark & \checkmark & \textbf{0.624} (\textbf{0.618}) & \textbf{0.481} (\textbf{0.477}) & \textbf{0.701} (\textbf{0.667}) & 0.751 (0.746) & \textbf{0.522} (\textbf{0.515}) & \textbf{0.604} (0.550) & \textbf{0.472} (\textbf{0.463})\\\hline
RVR & SD & MS & loudspeaker & rifle & sofa & table & telephone & watercraft & mean \\\hline\hline
& & & 0.584 (0.614) & 0.500 (0.576) & 0.624 (\textbf{0.643}) & 0.406 (0.427) & 0.522 (\textbf{0.705}) & 0.574 (\textbf{0.575}) & 0.516 (0.569)\\
\checkmark & & & 0.605 (0.598) & 0.570 (0.444) & 0.613 (0.612) & 0.431 (0.427) & 0.651 (0.590) & 0.570 (0.551) & 0.566 (0.536)\\
\checkmark & \checkmark & & 0.606 (0.598) & 0.565 (0.560) & \textbf{0.633} (0.632) & 0.429 (0.424) & 0.698 (0.638) & 0.550 (0.551) & 0.580 (0.565)\\
\checkmark & \checkmark & \checkmark & \textbf{0.641} (\textbf{0.624}) & \textbf{0.612} (\textbf{0.606}) & 0.622 (0.620) & \textbf{0.478} (\textbf{0.470}) & \textbf{0.719} (0.673) & \textbf{0.586} (0.569) & \textbf{0.601} (\textbf{0.584})\\\hline
\end{tabular}
\end{center}
\caption{Quantitative ablation studies for random view reinforced training (RVR), shape discriminator (SD) and multi-scale progressive training (MS). Numbers outside and inside parenthesis are mean voxel IoU scores on ShapeNet-Synthetic test set for ground truth and predicted viewpoints respectively. By adopting all three strategies, we can achieve the highest IoU scores in most cases with or without groud truth views.}
\label{tab:abalation_shapenet}
\end{table*}

\begin{figure*}
\begin{center}
\includegraphics[width=0.99\linewidth]{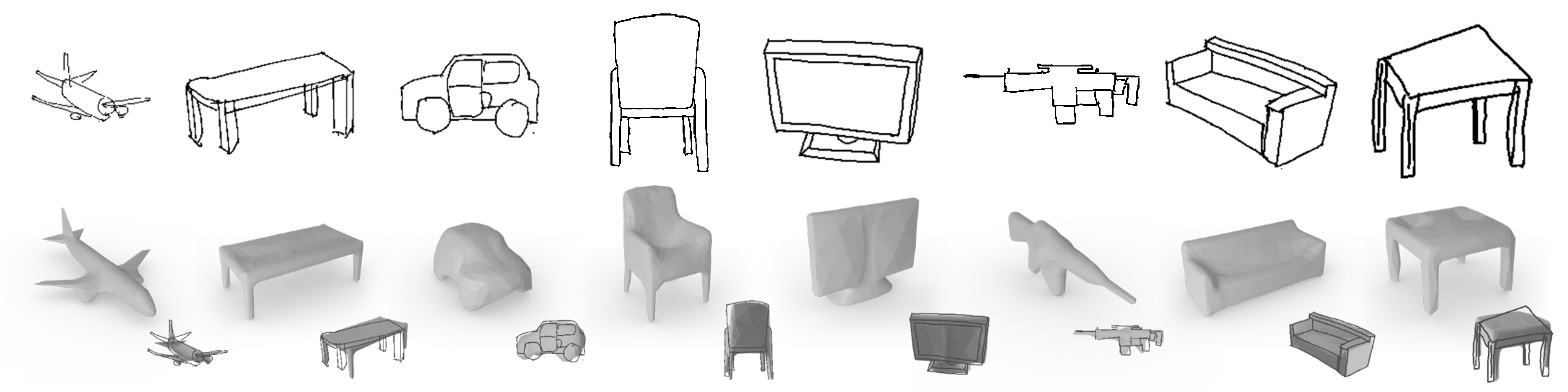}
\end{center}
   \caption{Some representative results on our ShapeNet-Sketch dataset. Each model is rendered from a fixed viewpoint, and the projection on the ground truth viewpoint is given on the bottom right corner of each example for comparison. Our method can generate promising shapes with hand-drawn sketches. Note that the results shown here use predicted viewpoints for generation. More results are shown in the supplementary material.}
\label{fig:ss}
\end{figure*}
\begin{table*}[h]
\footnotesize
\begin{center}
\renewcommand\tabcolsep{3.8pt}
\begin{tabular}{c||ccccccccccccc|c}
\hline
 Category & airplane & bench & cabinet & car & chair & display & lamp & loudspeaker & rifle & sofa & table & telephone & watercraft & mean \\\hline\hline
Retrieval & 0.411 & 0.219 & 0.409 & 0.626 & 0.238 & 0.338 & 0.223 & 0.365 & 0.413 & 0.431 & 0.232 & 0.536 & 0.369 & 0.370\\
SoftRas & 0.469  & 0.347  & 0.545  & 0.648  & 0.361  & 0.472  & 0.328  & 0.533  & \textbf{0.541}  & 0.534  & 0.359  & 0.537  & 0.456  & 0.472\\\hline
Ours (pred) & 0.479  & 0.357  & 0.547  & 0.649  & 0.383  & 0.435  & 0.336  & 0.526  & 0.510  & 0.528  & \textbf{0.361}  & 0.551  & 0.450  & 0.470\\
Ours (gt) & 0.487  & 0.366  & \textbf{0.568}  & 0.659  & 0.393  & \textbf{0.479}  & \textbf{0.338}  & \textbf{0.544}  & 0.534  & 0.534  & 0.357  & \textbf{0.554}  & \textbf{0.466}  & 0.483\\
Ours + DA (pred) & 0.515  & 0.362  & -  & 0.659  & 0.385  & -  & -  & -  & 0.511  & 0.533  & 0.360  & -  & -  & 0.475\\
Ours + DA (gt) & \textbf{0.526}  & \textbf{0.367}  & -  & \textbf{0.679}  & \textbf{0.398}  & -  & -  & -  & 0.535  & \textbf{0.548}  & 0.357  & -  & -  & \textbf{0.489}\\
\hline
\end{tabular}
\end{center}
\caption{Mean voxel IoU on ShapeNet-Sketch dataset. We apply domain adaptation (DA) on 7 of the classes, which have considerable amount of sketches in the Sketchy database~\cite{sangkloy2016sketchy} and Tu-Berlin dataset~\cite{eitz2012humans}. Our method with domain adaptation achieves the highest IoU score, and shows significant improvement on airplane, car and sofa.}
\label{tab:compare_shapenet_sketch}
\end{table*}

\begin{table*}[h]
\footnotesize
\begin{center}
\renewcommand\tabcolsep{4.3pt}
\begin{tabular}{c||ccccccccccccc||c}
\hline
 Category & airplane & bench & cabinet & car & chair & display & lamp & loudspeaker & rifle & sofa & table & telephone & watercraft & mean \\\hline\hline
SoftRas & 0.584 & 0.687 & 0.835 & 0.793 & 0.728 & 0.765 & 0.599 & 0.762 & 0.568 & 0.773 & 0.720 & 0.802 & 0.690 & 0.716\\\hline
Ours (pred) & \textbf{0.647} & 0.620 & 0.857 & 0.864 & 0.724 & 0.777 & \textbf{0.652} & 0.807 & \textbf{0.654} & 0.808 & 0.703 & 0.848 & 0.726 & 0.745\\
Ours (gt) & 0.644 & \textbf{0.673} & \textbf{0.868} & \textbf{0.866} & \textbf{0.737} & \textbf{0.807} & 0.631 & \textbf{0.822} & 0.614 & \textbf{0.817} & \textbf{0.735} & \textbf{0.855} & \textbf{0.735} & \textbf{0.754}\\
\hline
\end{tabular}
\end{center}
\caption{Comparison of 2D silhouette IoU score on ShapeNet-Sketch dataset. The generated shapes are projected to the ground truth view and we calculate the IoU score between the projected silhouette and the ground truth silhouette. Our method achieves significantly higher results than baseline methods. It demonstrates that our method can generate shapes better match the input sketches.}
\label{tab:iou_shapenet}
\end{table*}

\begin{figure*}
\begin{center}
\includegraphics[width=0.99\linewidth]{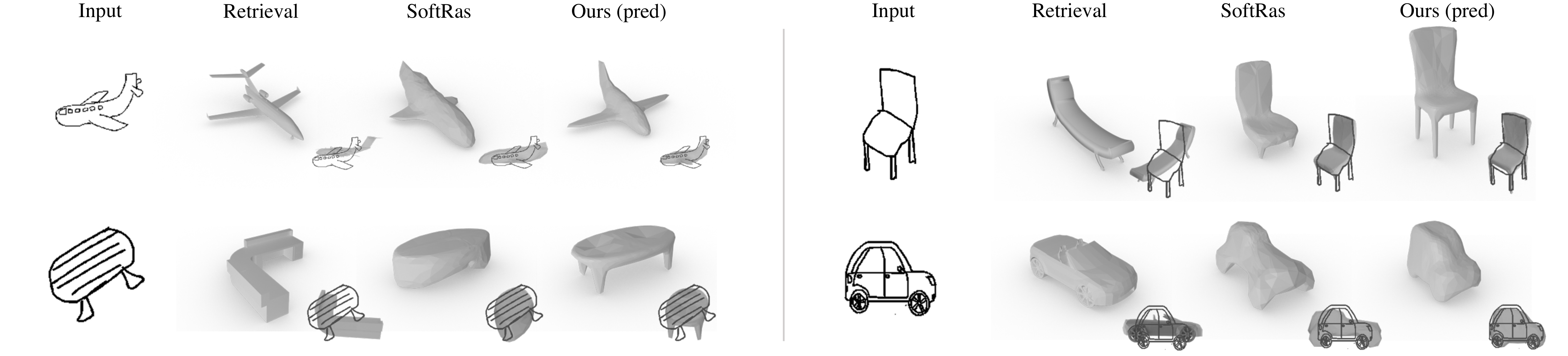}
\end{center}
   \caption{Qualitative comparisons with two baseline methods on sketches from the Sketchy database~\cite{sangkloy2016sketchy} and Tu-Berlin dataset~\cite{eitz2012humans}. The reconstructed mesh is projected on our predict viewpoint, to compare with the given sketch. Our method generates meshes that align better with the given sketch, and performs well on rare shapes (bottom right) and poorly-drawn sketches (bottom left).}
\label{fig:compare}
\end{figure*}
\begin{figure}
\begin{center}
\includegraphics[width=0.99\linewidth]{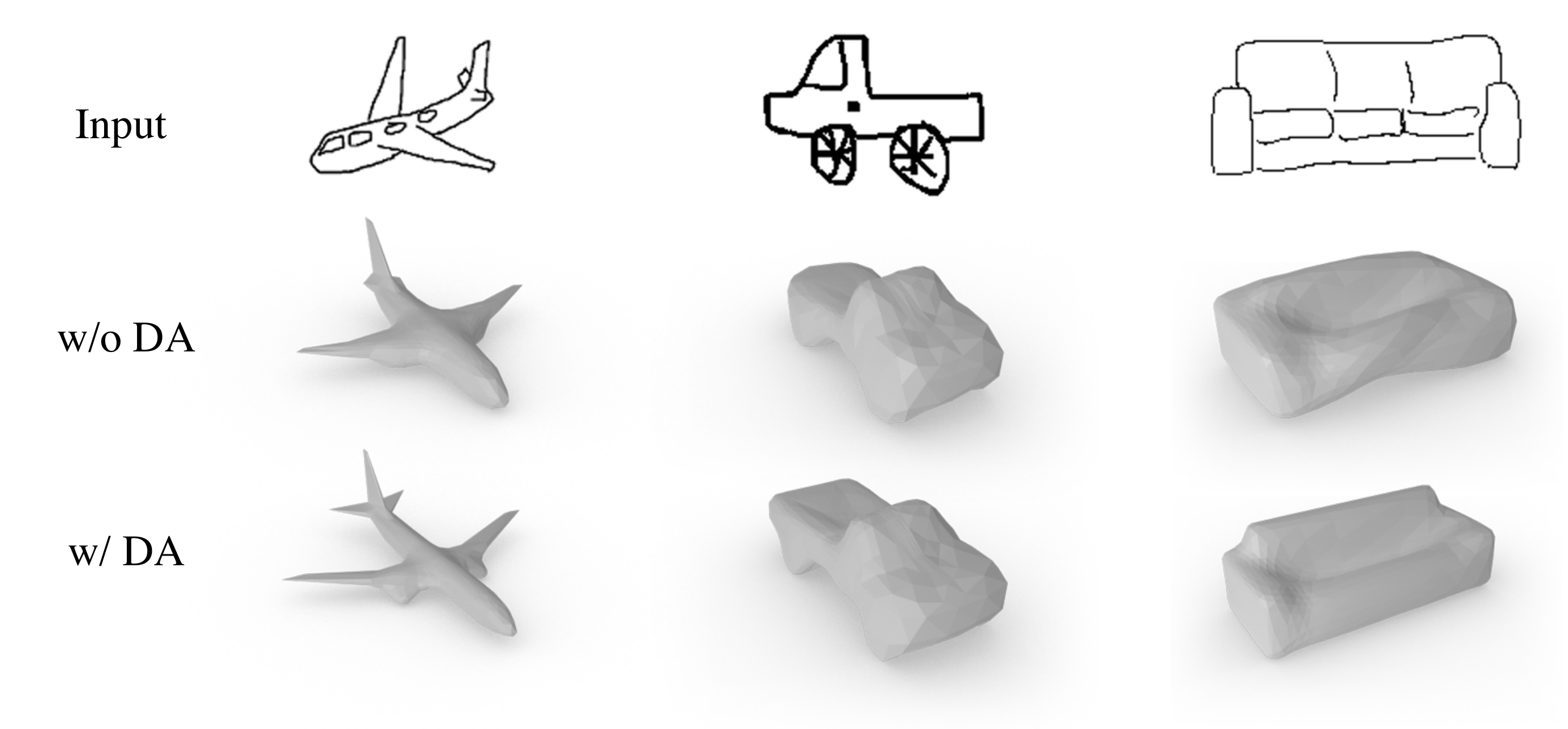}
\end{center}
   \caption{Ablation study for domain adaptation. Without domain adaptation, some distinctive features might be lost, like the tail of the airplane (left). Shapes generated after domain adaptation tend to have sharper boundaries (middle, right).}
\label{fig:compare_da}
\end{figure}

\section{Experiments}
We first perform case studies to show the effectiveness of our view-aware design and how specified viewpoints can be used to solve the ambiguity problem. Quantitative and qualitative evaluations on both synthetic and hand-drawn sketches show that our method generates shapes better matching the given sketch comparing to alternative baselines.
\subsection{Datasets}
\paragraph{ShapeNet-Synthetic.} We use synthetic sketches to train our model. We create the synthetic dataset using rendered images provided by Kar \etal~\cite{kar2017learning}. It contains object renderings of 13 categories from ShapeNet~\cite{chang2015shapenet}. Images from 20 different views of each object are rendered in 224x224 resolution. We extract the edge map of each rendered image using canny edge detector, which does not require 3D ground truth comparing to extracting contours from 3D shapes. We directly use the edge maps as synthetic sketches, and use this dataset for training and evaluation, under the same train/test split as in~\cite{liu2019soft}.
\paragraph{ShapeNet-Sketch.} To quantitatively evaluate our method on free-hand sketches and benefit further research, we collect a paired dataset of 3D shape and hand-drawn sketches, which we call ShapeNet-Sketch. We randomly choose 100 rendered images for each of the 13 ShapeNet categories from Kar's dataset~\cite{kar2017learning}, and ask people of different drawing skills to sketch the object based on the rendered image on a touch pad. Our ShapeNet-Sketch dataset contains in total 1,300 free-hand sketches and their corresponding ground truth 3D models. We use this dataset for evaluations only.
\paragraph{Existing sketch datasets.} 
We use some existing sketch datasets designed for other sketch-related tasks to qualitatively evaluate our method, and demonstrate the advantages of our method in dealing with poorly-drawn sketches. In detail, we use sketches of 7 categories that overlap with the ShapeNet classes, from Sketchy database~\cite{sangkloy2016sketchy} and Tu-Berlin dataset~\cite{eitz2012humans}.

\subsection{Implementation Details}
We use ResNet-18~\cite{he2016deep} as our image feature extractor and SoftRas~\cite{liu2019soft} for differentiable rendering of generated meshes. Detailed network architecture is described in the supplementary material. We assume a canonical view for objects in each class and use elevation and azimuth angles to represent viewpoints. The canonical view is 0 in elevation and 0 in azimuth, and the distance to camera is fixed. We use Adam optimizer with learning rate $1e-4$, and train a separate model for each class. Hyper-parameters are set to $\lambda_r=\lambda_{sd}=\lambda_{dd}=0.1$, $\lambda_v=\lambda_{vr}=10$.

\subsection{Effectiveness of View-Aware Design}
\label{sec:view-aware-exp}
As is illustrated in Sec.\ref{sec:view-aware}, our generation result relies on both input sketch image and a viewpoint. With different viewpoints, the network is expected to generate different shapes to match the input sketch. Examples in Fig.\ref{fig:view_aware} show how the generated shape changes with the change of elevation and azimuth angle. For each example, the second row shows how well the generated model matches the input sketch from the given view. In the vehicle example (top right), the choice of azimuth angle, which represents which side the vehicle is looking from, influences the output model to be a van (left, -90 in azimuth) or a truck (right, 90 in azimuth). For the badly-drawn table (bottom left), the height of the table and the thickness of the table top are determined by the elevation angle. 

Fig.\ref{fig:view_aware_ablation} shows the importance of the proposed random view reinforced training strategy and shape discriminator to the generation results. If only normal views are used in training, the network fails to learn the dependency on input viewpoint, and could get confused on given views. Without shape discriminator, the output mesh may get distorted and not resemble a real object. Under the influence of both, we can get view-aware property with promising shape quality.

\subsection{Comparisons and Evaluations}
We compare our method with the traditional encoder-decoder architecture in Fig.~\ref{fig:network}(a) and a retrieval-based approach. For the former, we adopt SoftRas~\cite{liu2019soft} as the differentiable rendering module, and denote it as ``SoftRas'' for simplicity. We also train a separate model for each category for fair comparison. Retrieval is a widely-adopted apporach to get 3D shape from a single sketch input. Instead of using traditional image features, we use features from a pre-trained sketch classification network, which is demonstrated to be more powerful than hand-crafted features in many sketch-related works~\cite{yu2015sketch,sangkloy2016sketchy}. For each input sketch, we find its nearest neighbor in training edge maps and take the corresponding 3D shape as the retrieval result. For quantitative comparisons, we use voxel IoU score, as in~\cite{kato2018neural,liu2019soft}.
\subsubsection{Comparisons on ShapeNet-Synthetic Dataset}
Comparison results on ShapeNet-Synthetic test set is shown in Table.\ref{tab:compare_shapenet}. We evaluate our method using both the predicted viewpoint (pred) and the ground truth viewpoint (gt), which serves as the lower and upper bound for our view-aware design. Our method outperforms the two baselines on most of the categories, especially when the ground truth view is given. This implies that view information plays an essential role in reconstruction, and generation quality should be improved by specifying correct views. We perform ablation studies on several designs of our approach in Table.\ref{tab:abalation_shapenet}. RVR, SD, MS denotes for random view reinforced training, shape discriminator and multi-scale progressive training respectively. Without using any of the three techniques, the method performs well on predicted viewpoints, but fails on given ground truth views, since it is not trained to fit arbitrary viewpoints. Adopting random view reinforced training without shape discriminator will lead to poor results on predicted views, because the shapes may get distorted given inaccurate view predictions. By applying multi-scale progressive training, the performance can be further improved. 
\subsubsection{Comparisons on ShapeNet-Sketch Dataset}
We show voxel IoU score on our collected ShapeNet-Sketch dataset in Table.\ref{tab:compare_shapenet_sketch}, and further demonstrate the effectiveness of our domain discriminator. We observe significant performance improvement after applying domain adaptation in some of the classes (airplane, car, sofa), indicating that large domain gaps exist. For comparison, we also tried to create fake sketches by utilizing CycleGAN and use them as training data, but no significant improvement is observed. Some reconstruction results using predicted views on ShapeNet-Sketch dataset is shown in Fig.\ref{fig:ss}. Fig.\ref{fig:compare_da} visualizes the effects of domain adaptation. Shapes without domain adaptation can be too smooth and lack distinctive features, like the tail of the airplane (left). 

We also compare projected silhouettes of generated meshes with ground truth silhouettes, and show the results in Table.\ref{tab:iou_shapenet}. It shows our model is more powerful at matching input sketches, which provides evidence for the effectiveness of our view-aware design. 
\subsubsection{Evaluations on Free-Hand Sketches}
Finally, we qualitatively evaluate our method on hand-drawn sketches from the Sketchy database~\cite{sangkloy2016sketchy} and Tu-Berlin dataset~\cite{eitz2012humans}. Some representative results are shown in Fig.\ref{fig:compare}. It can be seen that our method can better align the boundary of the generated mesh with the input sketch, benefit from the view-aware design. Our method balance well between sketch alignment and shape realism when handling out-of-distribution shapes, like the airplane on the top left and the table on the bottom left. Results here are all generated automatically using predicted viewpoints, and users should be able to control the output shape by specifying new viewpoints, like is illustrated in Sec.\ref{sec:view-aware-exp}. This makes our method more flexible and practical than existing view-agnostic methods.

\section{Conclusion}
We investigate the problem of generating 3D mesh from a single free-hand sketch, aiming for fast content generation for novice users. A view-aware architecture is proposed, to condition the generation process explicitly on viewpoints, which improves generation quality and brings controllability to the output shape. Our method can generate promising shapes on several free-hand sketch datasets, well balancing user intentions and shape qualities, especially for poorly-drawn sketches. We hope the view-aware paradigm can inspire further researches in single view 3D reconstruction and sketch-related areas.

\paragraph{Acknowledgements.} We thank all reviewers for their thoughtful comments. This work was supported by the National Key Technology R\&D Program (Project Number 2017YFB1002604), the National Natural Science Foundation of China (Project Numbers 61772298, 61521002), Research Grant of Beijing Higher Institution Engineering Research Center, and Tsinghua–Tencent Joint Laboratory for Internet Innovation Technology.

{\small
\bibliographystyle{ieee_fullname}
\bibliography{cvpr}
}

\end{document}